\documentclass{article} 
\usepackage{iclr2023_conference,times}
\usepackage{graphicx}
\usepackage{xcolor}
\usepackage{caption}
\usepackage{subcaption}
\usepackage{mwe}

\usepackage{amsmath,amsfonts,bm}

\def\eqref#1{equation~\ref{#1}}

\def\1{\bm{1}}

\DeclareMathAlphabet{\mathsfit}{\encodingdefault}{\sfdefault}{m}{sl}
\SetMathAlphabet{\mathsfit}{bold}{\encodingdefault}{\sfdefault}{bx}{n}

\usepackage{hyperref}
\usepackage{url}

\title{Accuracy is not the only Metric that matters: Estimating the
Energy Consumption of Deep Learning Models}

\author{%
  Johannes Getzner, Bertrand Charpentier\thanks{Corresponding author}, Stephan Günnemann \\
  Technical University of Munich, Germany\\
  \texttt{\{getzner,charpent,guennemann\}@in.tum.de} \\
}

\iclrfinalcopy 
\begin{document}

\maketitle

\begin{abstract}
Modern machine learning models have started to consume incredible amounts of energy, thus incurring large carbon footprints \citep{DBLP:journals/corr/abs-1906-02243}. To address this issue, we have created an energy estimation pipeline\footnote{\url{https://www.cs.cit.tum.de/daml/energy-consumption-dl}}, which allows practitioners to estimate the energy needs of their models in advance, without actually running or training them. We accomplished this, by collecting high-quality energy data and building a first baseline model, capable of predicting the energy consumption of DL models by accumulating their estimated layer-wise energies.
\end{abstract}
\section{Introduction}
\label{sec:introduction} 
Deep CNNs, such as VGG16 or ResNet50 already deliver great performance \citep{vgg_simonyan2014very,restnet_DBLP:journals/corr/HeZRS15}. Yet the increasing number of layers in such models comes at the cost of severely increased computational complexity, resulting in the need for power-hungry hardware \citep{comp_lim_DBLP:journals/corr/abs-2007-05558, comp_lim2_DBLP:journals/corr/JinYIK16}. An example of a model that behaves extremely poorly in this regard is a big transformer with neural architecture search \citep{DBLP:journals/corr/abs-1906-02243}. Clearly, training and running these models is not just a matter of financial cost, but also environmental impact. Yet, deep learning models are often only evaluated concerning standard performance metrics like accuracy or inference time, but not energy consumption or ecological footprint.  
While advancements in the research field often take precedence over energy considerations, this trade-off between performance and environmental impact should still be taken into account, despite the difficulty in accurately reporting these metrics due to a lack of appropriate tools \citep{impact_DBLP:journals/corr/abs-1910-09700, estimation_of_energy_GARCIAMARTIN201975}.

\looseness=-1
In light of these motivations, we developed a pipeline to collect model energy data and to approximate a model's energy consumption, before training and without the need for a direct energy measurement. Not only would this approach be more efficient and environmentally friendly, but it would also allow deep learning engineers to strike a balance between performance and environmental impact with greater ease. To achieve this goal, we attempt to compute the energy consumption of various deep learning architectures as the sum of their layer-wise estimated energies.

\textbf{Related work.} While previous publications in this field are rather scarce, some notable contributions helped tremendously with our research. The \emph{experiment-impact-tracker} is a python package that allows the user to measure their models and report their carbon footprint \citep{impact-tracker_DBLP:journals/corr/abs-2002-05651}. Another python package and integral part of our research is \emph{codecarbon}, which also allows for accurate CPU energy measurements and carbon reporting \citep{codecarbon}.  Additionally, \citet{impact_DBLP:journals/corr/abs-1910-09700} created \emph{ML CO2 Impact}, a web tool to calculate a models CO2 impact based on key parameters. Furthermore, in \citet{NeuralPower_DBLP:journals/corr/abs-1710-05420} the authors also  attempt to estimate the energy consumption of CNNs through layer-wise estimates. Unfortunately, their work shows some limitations as the energy consumption was measured from a GPU with a fixed clock speed, which is not realistic for real-world use. Moreover, the authors utilize hardware features in their models, making their tool more hardware-dependent. The authors of \citet{misnomer} further motivate our cause by promoting the usage of concrete real-world metrics, and show that no single cost indicator is sufficient and that incomplete cost reporting can have a direct environmental impact. \\

\looseness=-1
\textbf{Our Contribution.} \textbf{(1)} We release a high-quality data set on the energy consumption of various models and layer types, along with a novel and modular data-collection process, built to make our research hardware independent (Sec.~\ref{sec:data-collection}). \textbf{(2)} We created a collection of predictors for different layer types, forming a simple energy-prediction baseline for the energy consumption of multiple Deep Learning (DL) architectures (Sec.~\ref{sec:simple-baseline}). \textbf{(3)} We conducted a complex analysis of the predictive capabilities of various feature sets, offering valuable insights into the energy behavior of different architectures and layer types (Sec.~\ref{sec:results}).
\section{Data Collection of DL Energy Consumption}
\label{sec:data-collection}
There are three main steps to the data-collection process. \textbf{(1)} A module type is selected from the assortment of modules available in the implementation (see Tab.~\ref{tab:layer-params}). \textbf{(2)} All required parameters for this type (e.g. kernel-size) are sampled randomly from configurable ranges. \textbf{(3)} We use codecarbon to measure the CPU energy consumption by performing several forward-passes through the randomly configured module. For an overview of all modules, parameters, and their ranges, refer to Tab.~\ref{tab:layer-params}.

\looseness=-1
We differentiate between two categories of modules and corresponding data sets. The ``layer-wise'' data contains energy measurements for modules, which represent the layer types (e.g. Conv2d\footnote{\url{https://pytorch.org/docs/stable/nn.html}}), and will be used to train the predictors described in the next section. The ``model-wise'' data set contains the same measurements but for complete architectures (e.g. VGG16), and is used as a ground truth to evaluate the final model energy estimates. Additionally, this data contains measurements of all individual layers of the architecture to help evaluate the prediction results in Sec.~\ref{sec:results} at a higher level of detail. Furthermore, we also calculate the MAC (multiply-accumulate-operations) count, a measure of computational cost, for each module. To summarize, both data sets contain the corresponding module parameters and the MAC count as feature inputs, and the measured CPU energy of the module as the output label. See Tab.~\ref{tab:data-sets} for the sizes of the data sets. Further technical details regarding this process can be found in Sec.~\ref{sec:appendix-data-collection} and the data files can be located in the repository. Lastly, as the MAC count is a crucial feature, but outside the scope of this discussion, refer to Sec.~\ref{sec:appendix-performance-measures} for more information. 
\begin{table}[t]
    \caption{All possible parameter configurations for the data-collection process for each layer-type; the Activations encompass: ReLU, Sigmoid, Tanh, and Softmax; the Architectures encompass AlexNet \citep{alexnet_krizhevsky2017imagenet}, and VGG11/13/16.}
    \label{tab:layer-params}
    \begin{center}
        \begin{tabular}{l|lllll}
        \multicolumn{1}{c}{\bf Parameter}  & \multicolumn{1}{c}{\bf Conv2d} & \multicolumn{1}{c}{\bf MaxPool2d} & \multicolumn{1}{c}{\bf Linear} & \multicolumn{1}{c}{\bf Activations} & \multicolumn{1}{c}{\bf Architectures} \\
        \\ \hline \\
batch-size        & [1, 256] & [1, 256]  & [1, 512]  & [1, 512]    & [1, 256]      \\
image-size        & [4, 224] & [4, 224]  & /         & /           & 224           \\
kernel-size       & [1, 11]  & [1, 11]   & /         & /           & /             \\
in-channels/size  & [1, 512] & /          & [1, 5000] & [$5*10^4$, $5*10^6$]    & 3             \\
out-channels/size & [1, 512] & [1, 512]  & [1, 5000] & /           & /             \\
stride            & [1, 5]   & [1, 5]    & /         & /           & /             \\
padding           & [0, 3]   & [0, 3]    & /         & /           & / 
        \end{tabular}
    \end{center}
\end{table}
\section{A Simple Baseline to Estimate DL Energy Consumption}
\label{sec:simple-baseline}

\looseness=-1
To calculate the energy consumption of a complete architecture, the first step is to parse its configuration and extract all layers. Modules, such as AdaptiveAvgPooling and Dropout, with a negligible contribution to the total energy consumption (Fig.~\ref{fig:layer-wise-contribution}) are discarded. We then built a separate predictor for each layer type to estimate the energy consumption for each instance of that type of layer.

\looseness=-1
The input features of each predictor vary depending on the layer type, but they are generally based on the layer parameters listed in Tab.~\ref{tab:layer-params}. Furthermore, to help the model capture potential non-linear trends in the energy consumption, we may also consider the log transformation of each parameter as an additional feature \citep{NeuralPower_DBLP:journals/corr/abs-1710-05420}. Moreover, for certain models, we decided to include the MAC count as a feature, since it encodes the layer's parameters and is commonly used as a measure of runtime, implying that it could also serve as an accurate predictor of a model's energy consumption. As we do not expect any higher-order dependencies or strong non-linear relationships we chose to use Linear/Polynomial Regression models to predict the CPU energy consumption of each layer. Additionally, these models offer superior transparency and interpretablity thanks to their simplicity. Next, due to its small scale, a MinMaxScaler is always applied to the target variable ``cpu-energy''. Furthermore, we occasionally opted to use a StandardScaler to compensate for large differences in feature scales between layer parameters and the MAC count. All models and transforms were taken from scikit-learn \footnote{\url{https://scikit-learn.org/stable/modules/classes.html\#module-sklearn.preprocessing}}.

\looseness=-1
After all the models were trained, we used them to estimate the energy of each layer in the architecture. The total energy estimate for the full architecture is then found by summing the individual layer predictions. We made this design choice because we anticipate that the complexities of the individual layers will accumulate accordingly.

\section{Performance and Evaluation of Layer-Wise Energy Prediction Models}
\label{sec:results}

\looseness=-1
In the following, we will show how well the predictors for each layer type perform when trained using the \textbf{``layer-wise''} data set (Sec.~\ref{sec:data-collection}), which was divided into three sets: 70\% for training, 20\% for validation, and 10\% for testing. We evaluated models with different feature sets for each layer type (App.~\ref{sec:appendix-feature-sets-experiment}), but chose the best model according to the test $R^2$ score, MSE, and Max-Error. Additionally, we evaluated the 10-fold cross-validation average $R^2$ score and MSE. See Tab.~\ref{tab:layer-modles-performances} for the exact scores and both Fig.~\ref{fig:layer-scatterplots-layers} and Fig.~\ref{fig:layer-scatterplots-activations} for scatter plots.

\looseness=-1
\textbf{Activations.} While often numerous, activations only make up for a small fraction of the model energy consumption (see Fig.~\ref{fig:layer-wise-contribution}). The input features of the Sigmoid, Tanh, and Softmax models are the layer parameters with a polynomial transform of $d=2$ and interaction-only terms. The ReLU model was based solely on MAC count. The models, in the same order, achieved $R^2$ test scores of  0.9905, 0.9761, 0.9913, and 0.9812 respectively. \textbf{Linear.} Compared to activations, Linear layers contribute a significant amount to the total energy needs (1\%-10\%). The predictor uses the MAC count as the only feature and achieved an $R^2$ test score of 0.9992. Other models, based on different feature sets, also performed exceptionally well, but we chose this one for its simplicity. \textbf{MaxPooling.} About as energy-intensive as the Linear layer, the MaxPooling layer is, due to the larger number of possible parameters, more difficult to model. The chosen model uses both standard and log-transformed layer parameters, the MAC count, and polynomial features of $d=2$ to achieve an $R^2$ test score of 0.9995. \textbf{Convolution.} Not only is the Convolutional layer the most energy-expensive of the four (80\%-90\%), but it also has the most configurable parameters, making it the toughest layer to model. Yet surprisingly, the best model uses the MAC count as the only input feature and achieves an $R^2$ test score of 0.9977. App.~\ref{sec:appendix-ablation-analysis} further outlines the importance of the MAC count for this layer.

\begin{figure}[h]
    \centering
    \includegraphics[width=0.85\textwidth]{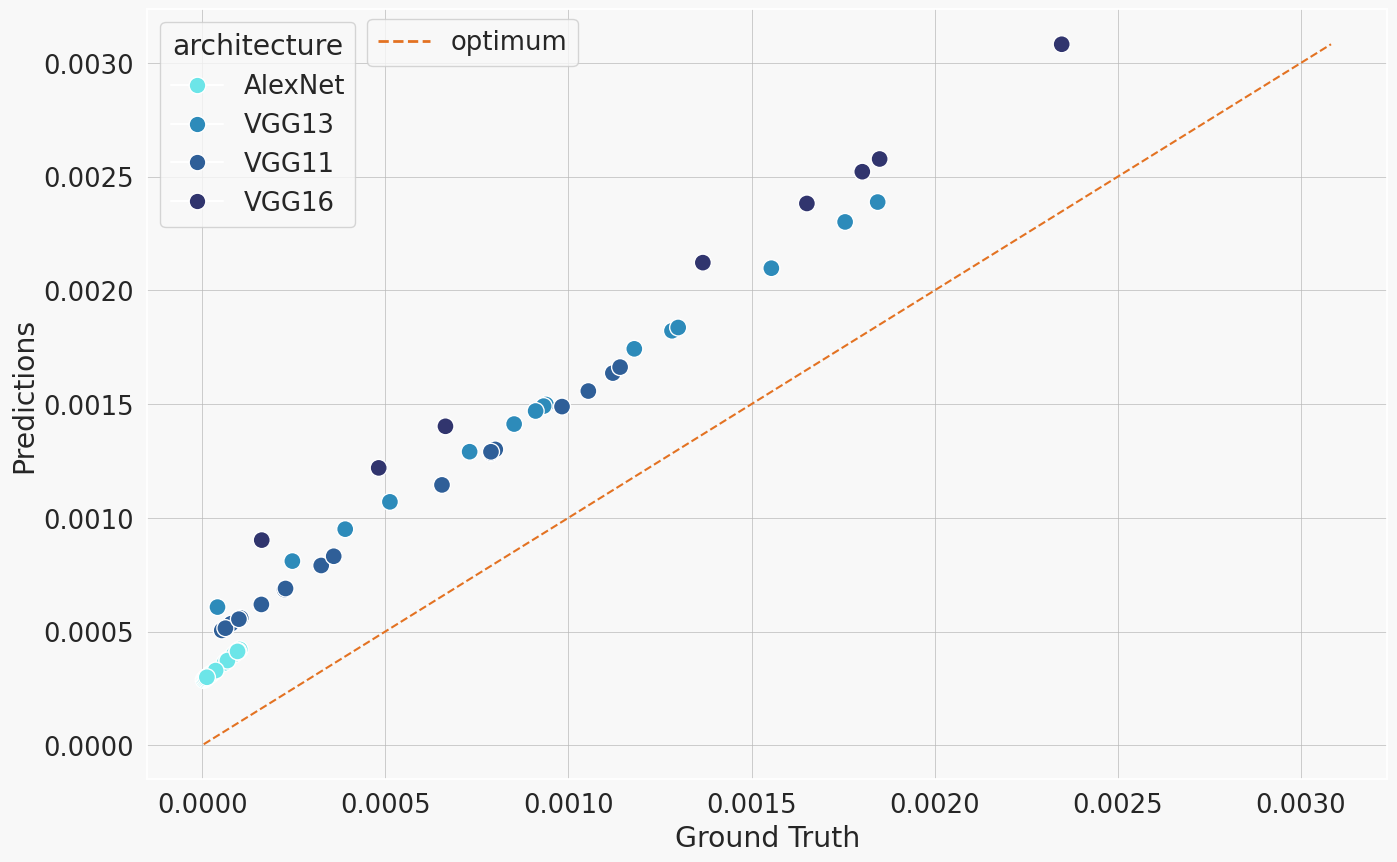}
    \caption{Compares the measured energy consumption and the accumulated predicted energy consumption; the x-axis corresponds to the measured ground truth, the y-axis to the sum of the layer-wise predictions, and the diagonal to the optimal scenario with perfect predictions; overall $R^2$ score: 0.352.}
    \label{fig:full-predictions}
\end{figure}

\begin{figure}[h]
        \centering
        \begin{subfigure}[b]{0.47\textwidth}
            \centering
            \includegraphics[width=\textwidth]{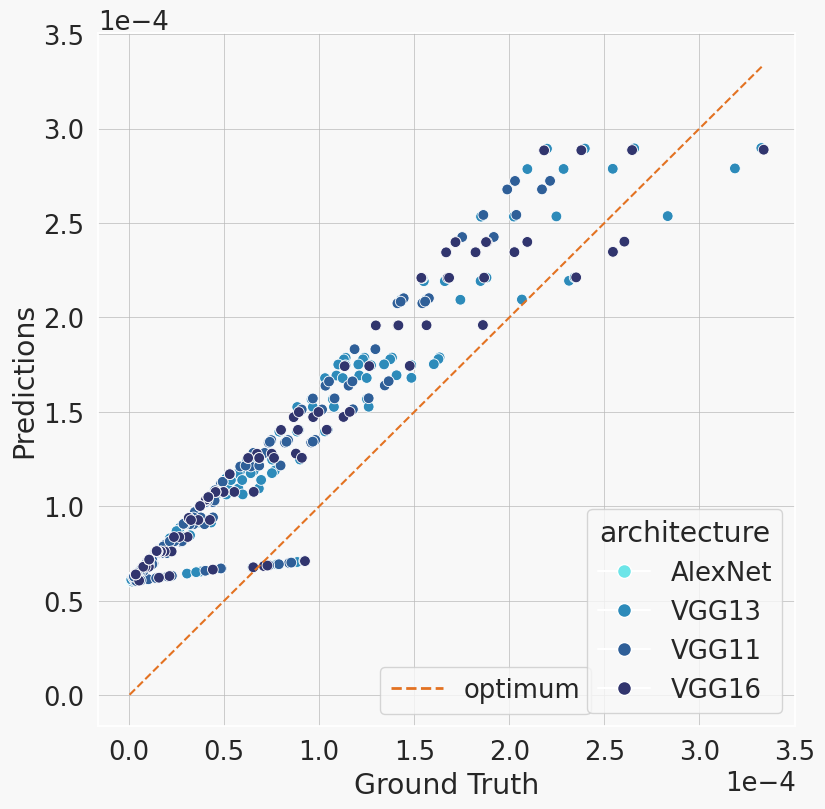}
            \caption[]%
            {{\small Convolution}}    
            \label{}
        \end{subfigure}
        \hfill
        \begin{subfigure}[b]{0.47\textwidth}  
            \centering 
            \includegraphics[width=\textwidth]{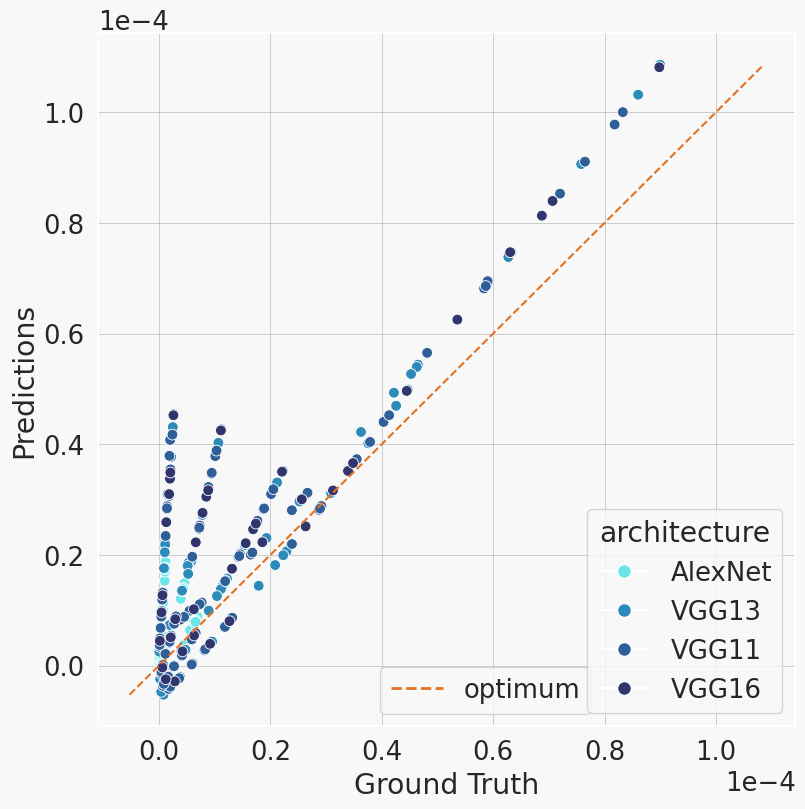}
            \caption[]%
            {{\small MaxPooling}}    
            \label{}
        \end{subfigure}
        \vskip\baselineskip
        \begin{subfigure}[b]{0.47\textwidth}   
            \centering 
            \includegraphics[width=\textwidth]{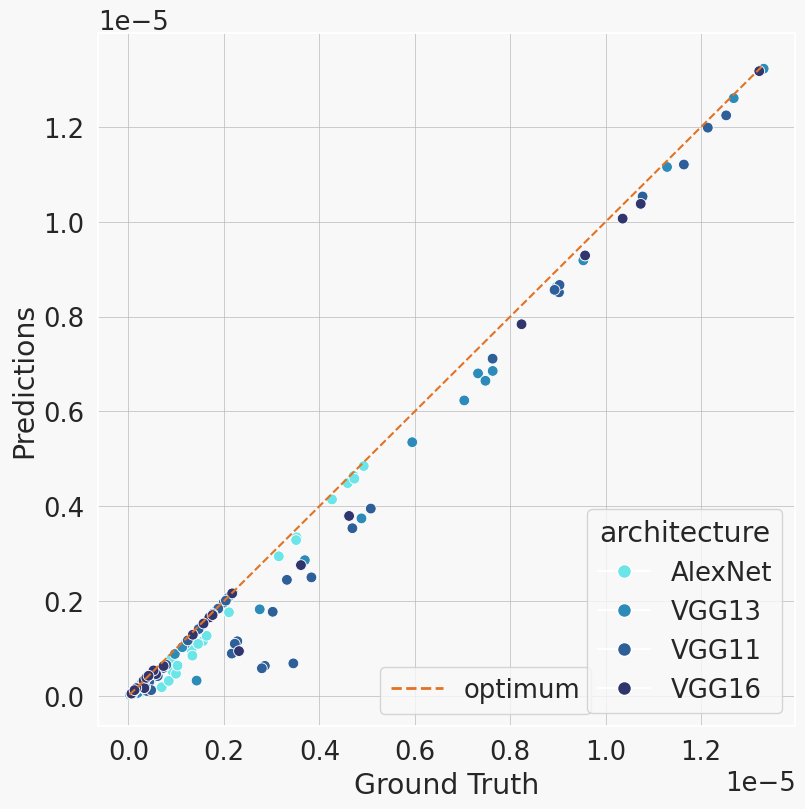}
            \caption[]%
            {{\small Linear}}    
            \label{}
        \end{subfigure}
        \hfill
        \begin{subfigure}[b]{0.47\textwidth}   
            \centering 
            \includegraphics[width=\textwidth]{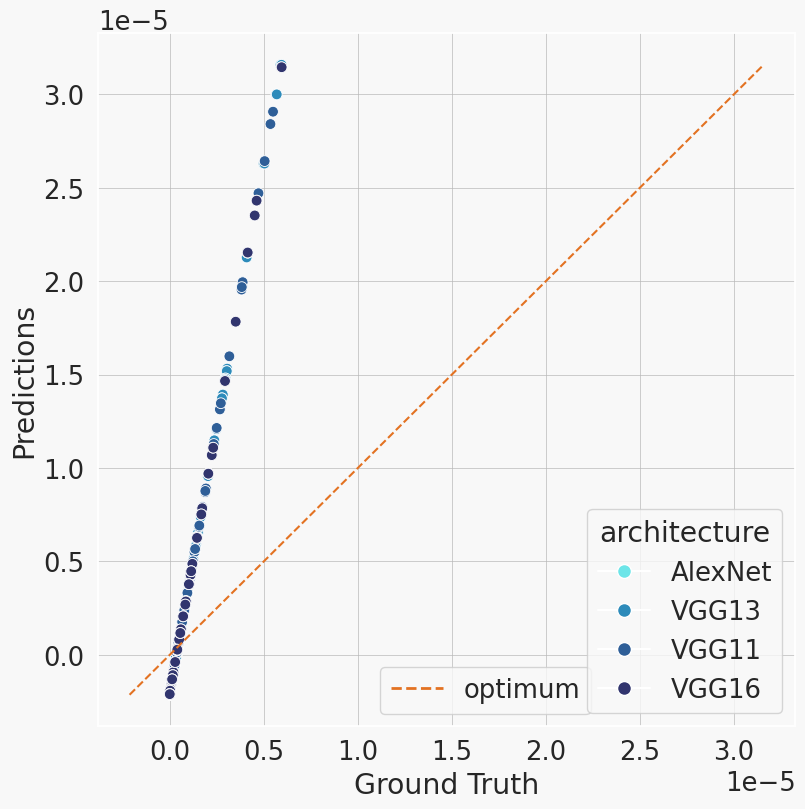}
            \caption[]%
            {{\small ReLU}}    
            \label{}
        \end{subfigure}
        \caption[Shows scatter plots for the layer types in AlexNet and VGG11/13/16, illustrating each model's accuracy when tested on real configurations from these architectures; the x-axis represents the ground truth while the y-axis represents the predicted values, and the diagonal line symbolizes perfect predictions; $R^2$ scores: 0.314 Convolution, 0.559 MaxPooling, 0.977 Linear, -21.51 ReLU.]
        {Shows scatter plots for the layer types in AlexNet and VGG11/13/16, illustrating each model's accuracy when tested on real configurations from these architectures; the x-axis represents the ground truth while the y-axis represents the predicted values, and the diagonal line symbolizes perfect predictions; $R^2$ scores: 0.314 Convolution, 0.559 MaxPooling, 0.977 Linear, -21.51 ReLU.}
        \label{fig:layer-preds-real-architectures}
\end{figure}

Given these exceptionally good results, we proceed by testing each predictor's capabilities not on random but real configurations, using the per-layer observations from the \textbf{''model-wise``} data set. As can be seen in Fig.~\ref{fig:layer-preds-real-architectures} the models perform reasonably well: the estimates for the Linear layer are close to perfect ($R^2$ 0.977); the MaxPooling model ($R^2$ 0.559), as well as the Convolutional model ($R^2$ 0.314), overestimate the target considerably; the ReLU model struggles to extrapolate to the new configurations, but fortunately the contribution of these layers to the total energy is negligible. Finally, we sum up all layer-wise estimates for each full architecture measurement in the ``model-wise'' data to evaluate the accuracy of the entire process. The assumption that the sum of layer-wise energy equals total energy consumption holds with negligible differences. (see Fig.~\ref{fig:agg-vs-total-energy}). Fig.~\ref{fig:full-predictions} shows the results of this process. It can be seen that the energy consumption is generally overestimated, resulting in an overall $R^2$ score of 0.352. The scatterplot also indicates that computationally more expensive architectures generally suffer from greater overestimation. Although the models introduced in the previous paragraph showed excellent performance for randomly configured modules, it is evident that the generalization to carefully conceptualized modules from real architectures is not trivial. One reason for this may be the vast set of possible random configurations for the Convolution and MaxPooling layers, of which most are unlikely to appear in the real world. Further experiment showed that integrating even a small additional number of layer configurations derived from actual architectures can result in significant performance improvements (see Sec.~\ref{section:add-real-configs}). Another reason could be that the MAC count not always accurately represents energy consumption due to parallelization and memory optimizations \citep{Dissecting_DBLP:journals/corr/abs-2107-11949, misnomer}.
\begin{table}[t]
    \caption{Performance metrics of the final models that were chosen to model the energy consumption for each layer type.}
    \label{tab:layer-modles-performances}
    \begin{center}
        \begin{tabular}{l|lllll}
        \multicolumn{1}{c}{\bf Module} & \multicolumn{1}{c}{\bf Avg. $R^2$ Cross-Val}  & \multicolumn{1}{c}{\bf Avg. MSE Cross-Val} & \multicolumn{1}{c}{\bf $R^2$ Test Set} & \multicolumn{1}{c}{\bf MSE Test-Set}
        \\ \hline \\
\textbf{Conv2d}     & 0.994 (± 0.005) & -2.291e-05 (± 1.329e-05) & 0.9977 & 2.779e-05  \\
\textbf{MaxPool2d}  & 0.999 (± 0.000) & -2.552e-06 (± 4.612e-06) & 0.9995 & 7.736e-07  \\
\textbf{Linear}     & 0.999 (± 0.000) & -4.284e-05 (± 1.425e-05) & 0.9992 & 3.384e-05  \\
\textbf{ReLU}       & 0.981 (± 0.005) & -1.046e-03 (± 2.284e-04) & 0.9812 & 8.998e-04  \\
\textbf{Sigmoid}    & 0.981 (± 0.008) & -1.047e-03 (± 1.866e-04) & 0.9905 & 7.538e-04  \\
\textbf{Tanh}       & 0.976 (± 0.008) & -1.315e-03 (± 4.252e-04) & 0.9761 & 1.412e-03  \\
\textbf{Softmax}    & 0.989 (± 0.004) & -5.671e-04 (± 1.599e-04) & 0.9913 & 4.972e-04
        \end{tabular}
    \end{center}
\end{table}
\section{Limitations}
\label{sec:limitations}
While we introduced a first approach to estimate the energy consumption of a full model without running it, our pipeline still has some limitations. First, codecarbon provides no real process isolation, meaning that other processes on the system could introduce noise to the measurements. Secondly, not all layer types (e.g. AdaptiveAvgPooling) are modeled in our solution but additional ones can easily be added due to the modular design of the framework. Finally, the focus of this work was on estimating CPU energy consumption, even though measuring the energy consumption of GPUs, which are widely used in deep learning applications, would also be relevant.
\section{Conclusion}
\label{sec:conclusion}
In this work, we introduced a novel open-source framework for predicting the energy consumption of DL architectures, to allow for a trade-off between performance, energy consumption, and consequently environmental impact. This includes an extensive and modular data-collection process, to facilitate the creation of more DL energy data sets. Furthermore, we propose a first energy prediction baseline, offering insights into the energy behavior of various layer types and show that the MAC count is an important feature for energy prediction. Furthermore, we demonstrate that while it is possible to predict the energy consumption of layers with random configurations, accurately generalizing to real-world architectures remains a significant challenge.
\newpage
\bibliography{iclr2023_conference}
\bibliographystyle{iclr2023_conference}
\newpage
\appendix
\section{Appendix}
\subsection{Data Collection: Technical Aspects \& Tools used}
\label{sec:appendix-data-collection}
The data-collection process was implemented in \emph{Python v3.9.12}. All implementations for the architectures such as VGG11 or individual layers such as the Convolution were taken from the \emph{PyTorch v1.11.0} framework with \emph{torchvision v0.12.0}. Furthermore, the package \emph{ptflops v0.6.9}\footnote{\url{https://github.com/sovrasov/flops-counter.pytorch}} was used to compute MAC counts, except for the MaxPool2d layer (see App.~\ref{sec:appendix-macs-computation}. Finally, the tool \emph{codecarbon v2.1.3}\footnote{\url{https://github.com/mlco2/codecarbon}}  was used to measure the energy consumption of each model and layer configuration. codecarbon accesses the Intel RAPL interface to read out the energy consumption of the CPU during the execution of a program. On the hardware side, the same machine of a slurm cluster with an Intel(R) Xeon(R) CPU E5-2630 v4 @ 2.20GHz and 256GB of memory was used for the entire process. The data collection was performed solely on the CPU and no GPU was employed. This decision was based on the common practice of using CPUs to run models at inference, and as \citet{misnomer} stated that 80-90\% of machine learning workload is dedicated to inference processing rather than training.

To reduce the noise and errors introduced by codecarbon, we measured the energy consumption by tracking as many forward passes through a module, as possible within a 30-second window. We then normalized this measurement by the number of forward passes, resulting in an average energy estimate for a single forward pass. To obtain even more precise measurements, we repeated this process up to three times per configuration and took the average of the results. It should also be mentioned that on rare occasions codecarbon logged an error while measuring, but because we collected at least three data points per configuration, we were able to remove these erroneous observations, without losing information.
%
\subsection{Computer Performance Measures}
\label{sec:appendix-performance-measures}
The number of \emph{Floating Point Operations}, in short \emph{FLOPs}, is a performance measure in computing that is widely used to determine the capabilities of a computer. The metric is defined as the total number of basic machine operations (floating-point additions or multiplications) that need to be executed by a program. Floating point operations often have a higher latency than other types of operations, which can greatly impact a program's overall execution time. This makes them a useful proxy for determining said execution time \citep{Dissecting_DBLP:journals/corr/abs-2107-11949}.

In this work, we are interested in the number of FLOPs that are required to compute a single forward pass through a deep CNN such as VGG11, as well as for all the individual layers/operations in the architecture. In deep learning, the type of operation that often dominates the number of FLOPs is the matrix-matrix product. Thus, the deeper the architecture, the more matrix-matrix products need to be computed, which results in a proportional increase in the number of FLOPs \citep{Dissecting_DBLP:journals/corr/abs-2107-11949}. Since the number of FLOPs can be seen as a proxy to the model's execution time, it can also be seen as a proxy for the energy consumption of the same model (on the given hardware)\citep{Energy-based_DBLP:journals/corr/abs-1808-00286}. Another popular and very similar metric is the MAC count (multiply-accumulate operations), which is better suited to represent operations such as matrix-matrix products. A single MAC is defined as the product of two numbers where the result of said product is then added to an accumulator. Modern systems are often capable of computing a MAC as a single operation, thus making the computation of matrix-matrix products more efficient. An approximate translation from FLOPs to MACs can be achieved by simply dividing the number of FLOPs by two. Because matrix-matrix products are incredibly important to the domain of deep learning and, as already mentioned, dominate the computational complexity of current models, we only refer to the MAC count in this work and translate from FLOPs to MACs where necessary.
\subsubsection{Computation of MAC count for each layer type}
\label{sec:appendix-macs-computation}
In the following, the computation of the number of MACs required for a forward pass through a variety of layer types will be discussed. Generally, the number of MACs for a CNN such as VGG11 is computed as the sum of layer-wise MAC counts. Given that the Convolution is an important layer of any CNN architecture, we will begin by introducing the MACs formula for this layer type. Note the following terminology: the input/output width of a Convolution will be referred to by $w_{in}/w_{out}$ and the input/output height by $h_{in}/h_{out}$. Accordingly, the number of input/output channels will be given by $c_{in}/c_{out}$. Furthermore, $k$ will correspond to the size of a symmetric kernel, $P$ to the padding parameter, and $B$ to the batch-size. Given this, the MAC count for a single forward pass through a Convolutional layer is given by:
$$\#MACs(Convolution) = k^2 * w_{out} * h_{out} * c_{in} * c_{out} * B$$
Tab.~\ref{tab:mac-count-formulas} shows the formulas for all other relevant components and layer types mentioned in this work. Note that both the ReLU activation and the MaxPooling layer do not require any multiply-accumulate operations, which is why we divide the number of FLOPs by two to approximate the MAC count.
\begin{table}[t]
    \caption{MAC count formulas for all relevant components and functions required in paper; the third column gives sum of the MACs for all instances of the given layer type for a single forward pass through VGG11 with $B=1$.}
    \label{tab:mac-count-formulas}
    \begin{center}
        \begin{tabular}{l|lllll}
        \multicolumn{1}{l}{\bf Module}  & \multicolumn{1}{l}{\bf MAC Formula} & \multicolumn{1}{l}{\bf VGG11 Example}
\\ \hline \\
\textbf{Convolution}    & $k^2 * w_{out} * h_{out} * c_{in} * c_{out} * B$      & 7492882432 \\
\textbf{Linear}& $w_{in} * h_{in} * c_{in} * c_{out} * B$              & 123642856 \\
\textbf{MaxPooling}     & ($k^2 * w_{out} * h_{out} * c_{in} * B)/2$            & 3060736 \\
\textbf{ReLU}           & $(w_{in} * h_{in} * c_{in} * B)/2$                    & 3717120 \\
\end{tabular}
    \end{center}
\end{table}
%
\subsection{Adding real layer configurations to the training sets}
\label{section:add-real-configs}
The results presented in Sec.~\ref{sec:results} showed that the models for the Convolution and MaxPooling layers were struggling to generalize to configurations from real architectures. One hypothesis was that the random configurations used in the training sets may not be adequate representations of the layers in real-world architectures. This could be because the large number of parameters in these layers leads to a wide range of potential random configurations that are very unlikely to appear in any carefully conceptualize architecture. Thus, in the following experiment, we additionally measured the energy consumption of each individual configuration of Linear, Convolution, and MaxPooling layers present in AlexNet and VGG11/13/16 with a random batch-size and added these measurements to the training sets. After retraining, all models showed slight performance improvements when evaluated on the same task as in Sec.~\ref{sec:results}, resulting in an overall $R^2$ score of 0.45 (previously 0.352). The exact performance improvements for each layer can be seen in Tab.~\ref{tab:improvements-from-real-configs}. These results indicate that enriching the training set with \emph{individual} configurations similar to the ones present in larger models can improve the energy predictions on the \emph{full} models.

\begin{table*}[h]
    \caption{$R^2$ score performance improvements on layers present in \emph{full} architectures after adding measurements from \emph{individual} of the Convolution, MaxPooling, and Linear layer configurations present in the full architectures to the training data sets  models.}
    \label{tab:improvements-from-real-configs}
    \centering
    \begin{tabular}{l|ll}
        \multicolumn{1}{l}{\bf Module}  & \multicolumn{1}{l}{\bf score before}  & \multicolumn{1}{c}{\bf score after}
        \\ \hline \\
        \textbf{MaxPool2d} & 0.559 & 0.679 (+ 0.120) \\
        \textbf{Conv2d} & 0.314  & 0.395 (+ 0.081)\\
        \textbf{Linear} & 0.977 & 0.978 (+ 0.001) \\
    \end{tabular}
\end{table*}
\subsection{Predictive capabilities of various feature sets}
\label{sec:appendix-feature-sets-experiment}
To find the best possible model for each layer type, we experimented with a variety of different feature combinations, and a separate model was fitted and fine-tuned to each, trying to maximize the performance in that setting. Thus, in the following, we will refer to the \emph{parameter-set} as a feature set that only contains the standard layer parameters, the \emph{(log+)parameter-set}, which in addition to the standard layer parameters also contains the log-transformed parameters, the \emph{parameter-MAC-set} as the feature set that contains both the layer parameters and the MAC count as features and finally the \emph{(log+)parameter-MAC-set} which contains all of the previously mentioned features. Additionally, we attempted to model each layer with the MAC count as the only feature. For all feature sets that contain the MAC count, we applied a StandardScaler before fitting the model to compensate for the large differences in magnitude between the parameters and the MAC count. The goal of this experiment was to analyze which feature combination would have the best predictive qualities for each layer. This analysis was not conducted for the activation functions.  For more details on the performance of all models, features-sets, and layer types please see Tab.~\ref{tab:feature-set-performances}.

\textbf{Linear layer}. All models are Linear regressors with the parameter- and (log+)parameters-set models having additional interaction-only polynomial features of degree $d=3$. Each one of the five configurations achieved an outstanding $R^2$ score of at least 0.999, with the model based on the parameter-set scoring the lowest ($R^2$ 0.999). To help with feature selection, a Lasso regressor was also used with the (log+)parameters-set, but showed no improvements over a standard Linear regressor. 

\textbf{Convolutional layer}. The models based on the parameter- and (log+)parameter-sets were Lasso regressors with additional interaction-only polynomials of $d=3$ and $d=4$. Even though the Lasso models proved to be slightly more robust than standard Linear regressors, the models performed poorly with the former achieving an $R^2$ score of 0.7097 and the latter only performing slightly better with a score of 0.7954. The MAC-count-only Linear regressor showed a significant boost in performance with an $R^2$ score of 0.9977. Further small improvements could be achieved by combining the (log+)parameter-sets with the MAC count in a Linear regressor for a $R^2$ score of 0.9979. Sec.~\ref{sec:appendix-ablation-analysis} of the Appendix gives further insights on feature importances through an ablation analysis. 

\textbf{MaxPooling layer}. Similar to before, we trained two Lasso regressors based on the parameter- and (log+)parameter-sets, with additional interaction-only polynomial features of $d=4$ and $d=3$. While the former performed poorly with an $R^2$ score of only 0.4468, the latter with log-features performed significantly better with an $R^2$ score of 0.8194. Even though further increasing the degree of the polynomials resulted in slightly better scores, the models were less robust. In accordance with previous results, the feature sets, which contained the MAC count caused a significant uplift in performance, with the (log+)parameter-MAC- and parameter-MAC-set performing almost identically with a $R^2$ score of $0.9995$ and $0.9967$. The MAC count only model performed the worst of the three with an $R^2$ score of $0.9797$.

The goal of this experiment was to analyze the predictive capabilities of the various feature sets for the different layer types. Each model for each layer-feature-set combination was carefully fitted and evaluated to achieve the best possible performance. The experiment has shown that A) the MAC count feature is vital when it comes to predicting the energy consumption and may even be sufficient to do so, B) polynomial features are helpful when modeling energy consumption, and C) while Lasso regressors can help with feature selection, they do not provide significant improvements in these experiments.
%
\subsection{Ablation analysis: Convolutional Layer}
\label{sec:appendix-ablation-analysis}
In the following experiment, we conducted an ablation analysis with the Convolutional layer model. The model is based on the (log+)parameter-MAC-set, which contains all the possible features for this layer (see App.~\ref{sec:appendix-feature-sets-experiment} for details). The model is a Linear regressor with a StandardScaler applied to the features before fitting. The experiment was conducted as follows. For each one of the 32767 possible feature subsets of minimum size one, we fitted the model and reported the test scores and errors. The goal of the experiment was to see, which features are the most important and which exact combination of features would lead to the best results. No polynomial features were included in this analysis as that would cause a combinatorial explosion.

\begin{figure}[t!]
    \centering
    \includegraphics[width=0.9\textwidth]{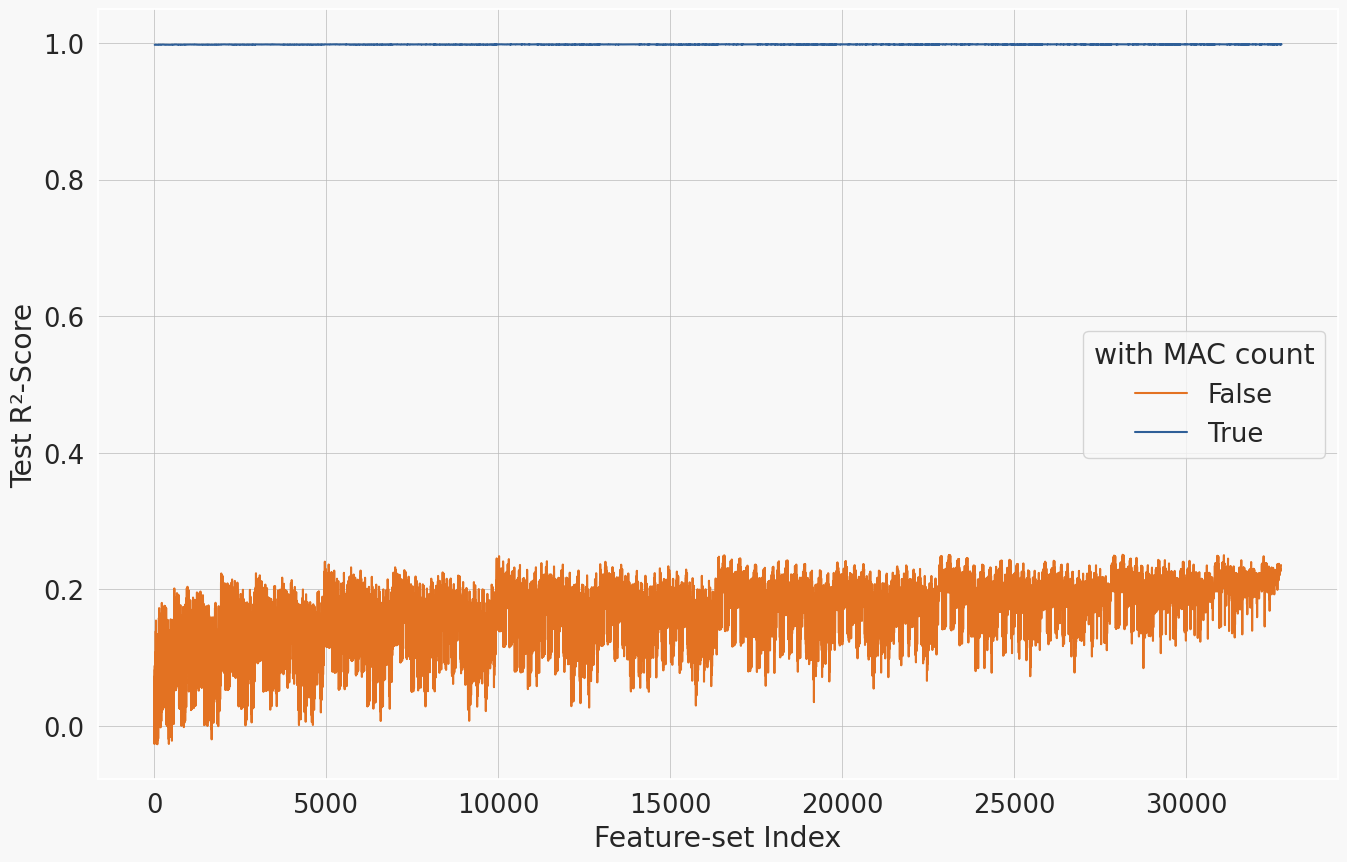}
    \caption{Shows the performance of all combinations of feature sets by their $R^2$ score; the x-axis corresponds to the feature set index and the y-axis to the $R^2$ score; the color indicates the whether the feature set contains the MAC count or not.}
    \label{fig:ablation-analysis}
\end{figure}

The results once again show that the MAC count is a very important predictor for the model. The maximum $R^2$ score for all models without this feature is only 0.25, while with the MAC count the score jumps to 0.998. Not only is the $R^2$ score with this feature substantially higher than without it, but also independent of the other features the score never reaches below 0.997. Furthermore, we observe that there are only very small performance deviations within the models that contain the MAC count. For all other feature sets, we observe much stronger changes in performance. Fig.~\ref{fig:ablation-analysis} illustrates this behavior and shows the performance of each feature set with and without the MAC count. Tab.~\ref{tab:ablation-analysis-results} shows the best and worst feature sets for both categories.
\begin{table*}[ht]
    \caption{Shows the best and worst feature combinations with their $R^2$ scores for both categories.}
    \label{tab:ablation-analysis-results}
    \centering
    \resizebox{1 \textwidth}{!}{
    \begin{tabular}{p{0.15\linewidth} p{0.42\linewidth} | p{0.42\linewidth}}
        \multicolumn{1}{c}{\bf Score}  & \multicolumn{1}{c}{\bf with MAC}  & \multicolumn{1}{c}{\bf without MAC}
        \\ \hline \\
        \textbf{highest score} & batch-size, image-size, kernel-size, in-channels, stride, log-image-size, log-kernel-size, log-out-channels, log-stride, log-padding, macs ($R^2$ $0.998$) & batch-size, image-size, kernel-size, in-channels, stride, log-batch-size, log-image-size, log-kernel-size, log-out-channels, log-stride ($R^2$ $0.25$) \\
        \\ \hline \\
        \textbf{lowest score}  & out-channels, log-out-channels, macs ($R^2$ $0.997$) & padding, log-padding ($R^2$ $-0.026$)\\
    \end{tabular}
    }
\end{table*}
%
\subsection{Additional Tables}
\begin{table}[h]
\caption{Shows the performance metrics for each layer type and feature set; the \emph{ito} in the \emph{Polynomial} column specifies that the polynomial features were restricted to interaction-only terms; the feature sets marked with a * were selected for the computation of the final estimates presented in Sec.~\ref{sec:results}}
\label{tab:feature-set-performances}
    \centering
    \resizebox{1 \textwidth}{!}{
    \begin{tabular}{lllcccccc}
            \multicolumn{1}{c}{\bf Module}  & \multicolumn{1}{c}{\bf Feature-Set}  & \multicolumn{1}{c}{\bf Polynomial} & \multicolumn{1}{c}{\bf StandardScaler} & \multicolumn{1}{c}{\bf Model} & \multicolumn{1}{c}{\bf Avg. $R^2$ Cross-Val} & \multicolumn{1}{c}{\bf Avg. MSE Cross-Val} & \multicolumn{1}{c}{\bf $R^2$ Test Set} & \multicolumn{1}{c}{\bf MSE Test-Set}
            \\ \hline \\
            \textbf{Conv2d}     & parameter             & d=4, ito & n & Lasso & 0.513 (± 0.293) & -2.467e-03 (± 1.889e-03) & 0.7097 & 3.490e-03  \\
                                & (log+)parameter       & d=3, ito & n & Lasso & 0.622 (± 0.192) & -2.105e-03 (± 1.734e-03) & 0.7954 & 2.460e-03  \\
                                & MACs*                  & & n & Linear & 0.994 (± 0.005) & -2.291e-05 (± 1.329e-5) & 0.9977 & 2.779e-05  \\
                                & parameter-MAC        & & y & Linear & 0.996 (± 0.004) & -1.644e-05 (± 9.509e-06) & 0.9979 & 2.501e-05  \\
                                & (log+)parameter-MAC  & & y & Linear & 0.996 (± 0.004) & -1.560e-05 (± 9.004e-06) & 0.9979 & 2.578e-05  \\
            \textbf{MaxPool2d}  & parameter             & d=4, ito & n & Lasso & 0.528 (± 0.185) & -1.375e-03 (± 1.941e-03) & 0.4468 & 9.338e-04  \\
                                & (log+)parameter       & d=3, ito & n & Lasso & 0.704 (± 0.139) & -1.139e-03 (± 2.137e-03) & 0.8194 & 3.048e-04  \\
                                & MACs                  & & n & Linear & 0.946 (± 0.051) & -9.746e-05 (± 7.278e-05) & 0.9797 & 3.431e-05  \\
                                & parameter-MAC        & d=2, ito & y & Linear & 0.993 (± 0.005) & -3.043e-05 (± 5.706e-05) & 0.9967 & 5.533e-06  \\
                                & (log+)parameter-MAC*  & d=2, ito & y & Linear & 0.999 (± 0.000) & -2.552e-06 (± 4.612e-06) & 0.9995 & 7.736e-07  \\
            \textbf{Linear}     & parameter             & d=3, ito &n& Linear & 0.999 (± 0.000) & -3.784e-05 (± 1.300e-05) & 0.9990 & 3.848e-05  \\
                                & (log+)parameter       & d=3, ito &n& Linear & 0.999 (± 0.000) & -2.756e-05 (± 1.179e-05) & 0.9995 & 1.839e-05  \\
                                & MACs*                  &&n& Linear & 0.999 (± 0.000) & -4.284-05 (± 1.425e-05) & 0.9992 & 3.384e-05  \\
                                & parameter-MAC        &&y& Linear & 0.999 (± 0.000) & -3.909e-05 (± 1.313e-05) & 0.9991 & 3.693e-05  \\
                                & (log+)parameter-MAC  &&y& Linear & 0.999 (± 0.000) & -3.466e-05 (± 1.147e-05) & 0.9992 & 3.134e-05  \\
            \textbf{ReLU}       & parameter             & d=2, ito & n & Linear & 0.981 (± 0.004) & -1.040e-03 (± 2.158e-04) & 0.9812 & 8.991e-04  \\
                                & MACs*                 & & n & Linear & 0.981 (± 0.005) & -1.046e-03 (± 2.284e-04) & 0.9812 & 1.030e-03  \\
            \textbf{Sidmoid}       & parameter*             & d=2, ito & n & Linear & 0.981 (± 0.008) & -1.047e-03 (± 1.866e-04) & 0.9905 & 7.538e-04  \\
            \textbf{Tanh}       & parameter*             & d=2 & n & Linear & 0.976 (± 0.008) & -1.315e-03 (± 4.252e-04) & 0.9761 & 1.412e-03  \\
            \textbf{Softmax}       & parameter*             & d=2, ito & n & Linear & 0.989 (± 0.004) & -5.671e-04 (± 1.599e-04) & 0.9913 & 4.972e-04  \\
                                
    \end{tabular}
    }
\end{table}
\begin{table}[h]
    \caption{Shows the number of unique and total data points that were collected for each PyTorch module; for the architectures, the number of data points also includes the data points for all individual layers, for each measurement of the full architecture.}
    \label{tab:data-sets}
    \begin{center}
        \begin{tabular}{llllll}
        \multicolumn{1}{c}{\bf Module} & \multicolumn{1}{c}{\bf Number of unique data points/configurations}  & \multicolumn{1}{c}{\bf Total data points}
\\ \hline \\
\textbf{Conv2d}     & 976 & 2900 \\
\textbf{MaxPool2d}  & 985 & 2939 \\
\textbf{Linear}     & 992 & 2962 \\
\textbf{ReLU} & 988 & 2951 \\
\textbf{Sigmoid} & 489 & 1467 \\
\textbf{Softmax} & 493 & 1479 \\
\textbf{Tanh} & 491 & 1473 \\
\textbf{AlexNet, VGG11/13/16} & 1840 (including layers) &  5801 (including layers)\\
\end{tabular}
    \end{center}
\end{table}
%
\newpage
\subsection{Additional Figures}
\begin{figure}[h]
        \centering
        \begin{subfigure}[b]{0.47\textwidth}
            \centering
            \includegraphics[width=\textwidth]{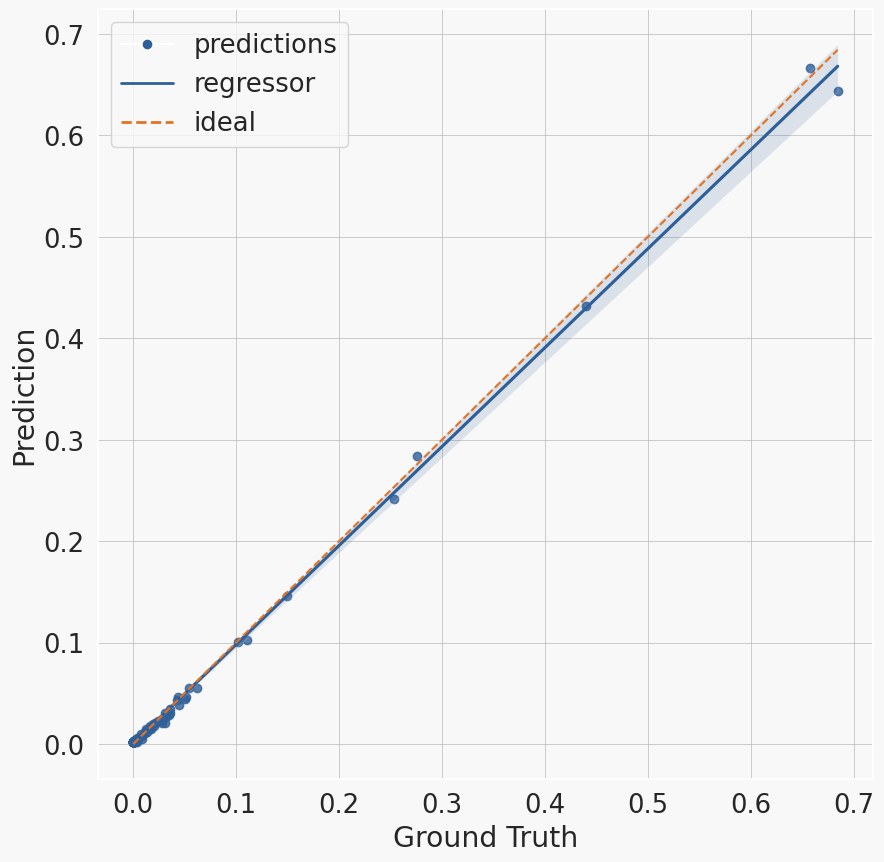}
            \caption[]%
            {{\small Convolution}}    
            \label{fig:layer-scatterplots convolution}
        \end{subfigure}
        \hfill
        \begin{subfigure}[b]{0.47\textwidth}  
            \centering 
            \includegraphics[width=\textwidth]{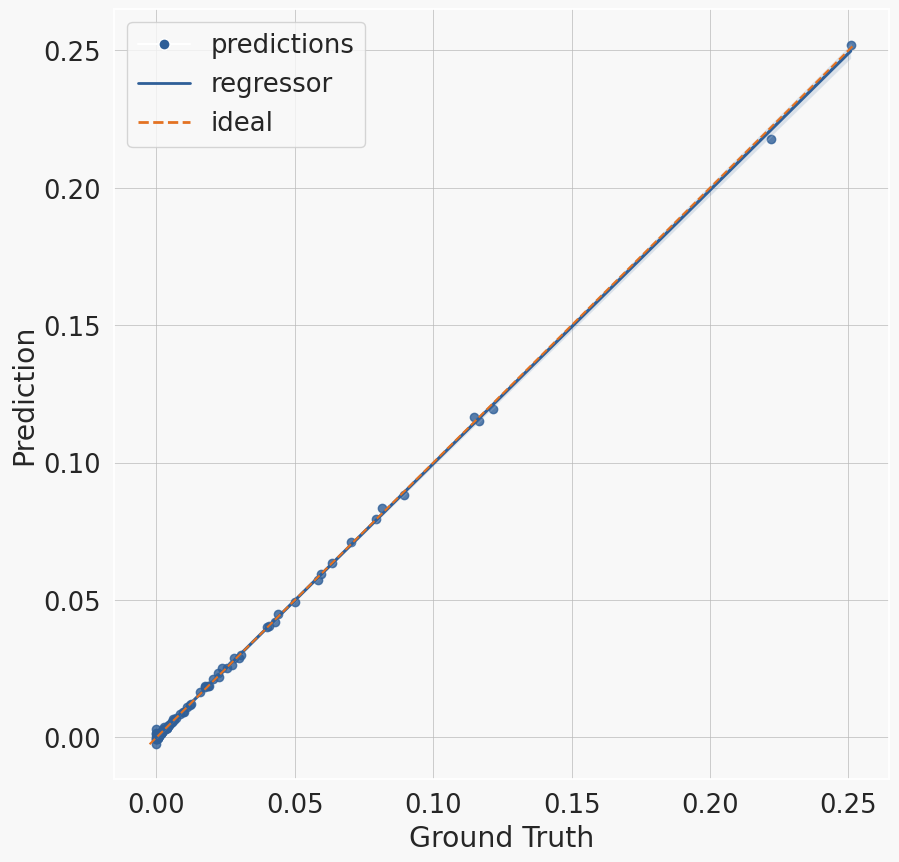}
            \caption[]%
            {{\small MaxPooling}}    
            \label{fig:layer-scatterplots maxpooling}
        \end{subfigure}
        \begin{subfigure}[b]{0.47\textwidth}   
            \centering 
            \includegraphics[width=\textwidth]{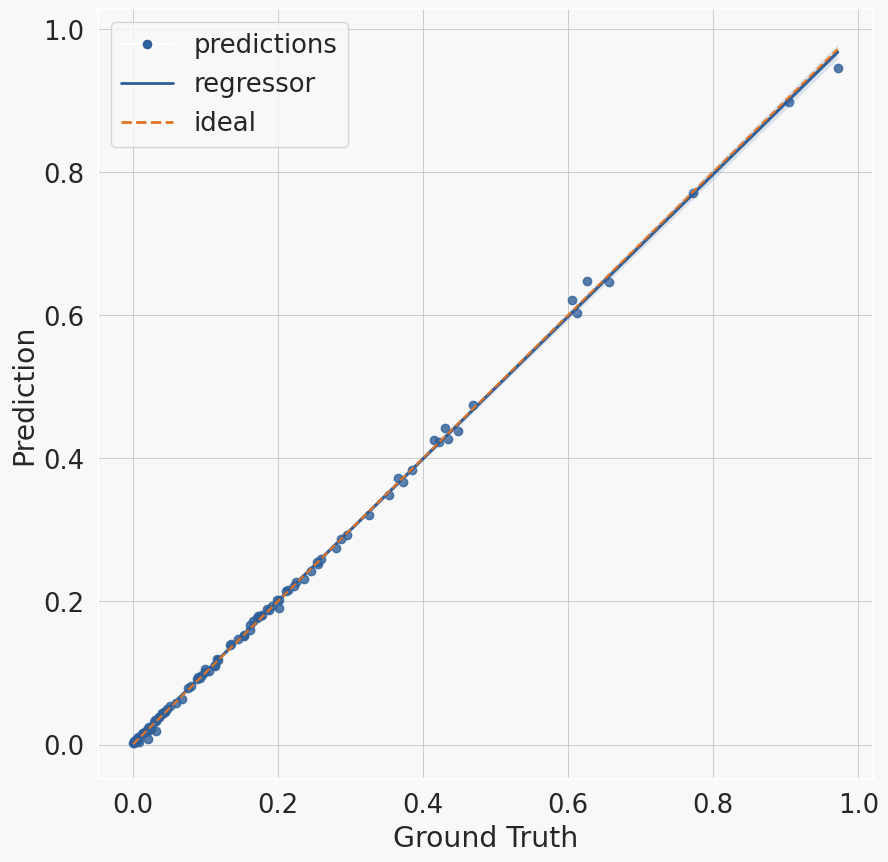}
            \caption[]%
            {{\small Linear}}    
            \label{fig:layer-scatterplots linear}
        \end{subfigure}
        \caption[Shows scatter-plots for the selected layer models (see Sec.~\ref{sec:results}), illustrating the accuracy of the predictions on the random test data; the x-axis represents the ground truth while the y-axis represents the predicted values, and the diagonal line symbolizes perfect predictions.]
        {Shows scatter-plots for the selected layer models (see Sec.~\ref{sec:results}), illustrating the accuracy of the predictions on the random test data; the x-axis represents the ground truth while the y-axis represents the predicted values, and the diagonal line symbolizes perfect predictions.}
        \label{fig:layer-scatterplots-layers}
\end{figure}
\begin{figure}[h]
        \centering
        \begin{subfigure}[b]{0.47\textwidth}
            \centering
            \includegraphics[width=\textwidth]{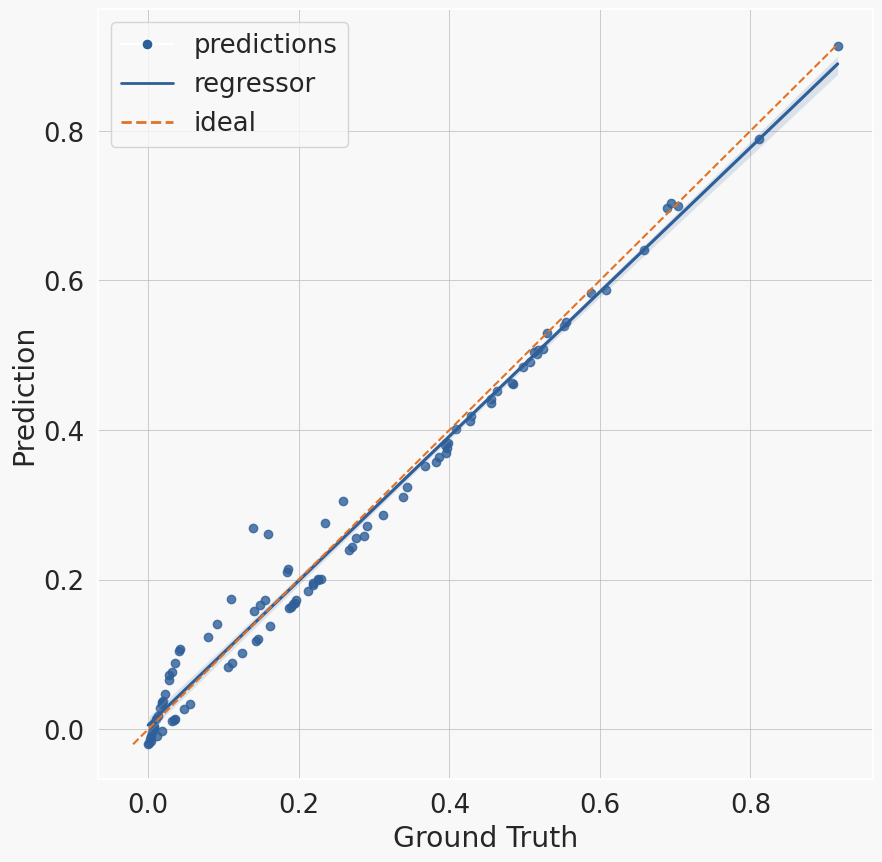}
            \caption[]%
            {{\small ReLU}}    
            \label{fig:layer-scatterplots relu}
        \end{subfigure}
        \hfill
        \begin{subfigure}[b]{0.47\textwidth}  
            \centering 
            \includegraphics[width=\textwidth]{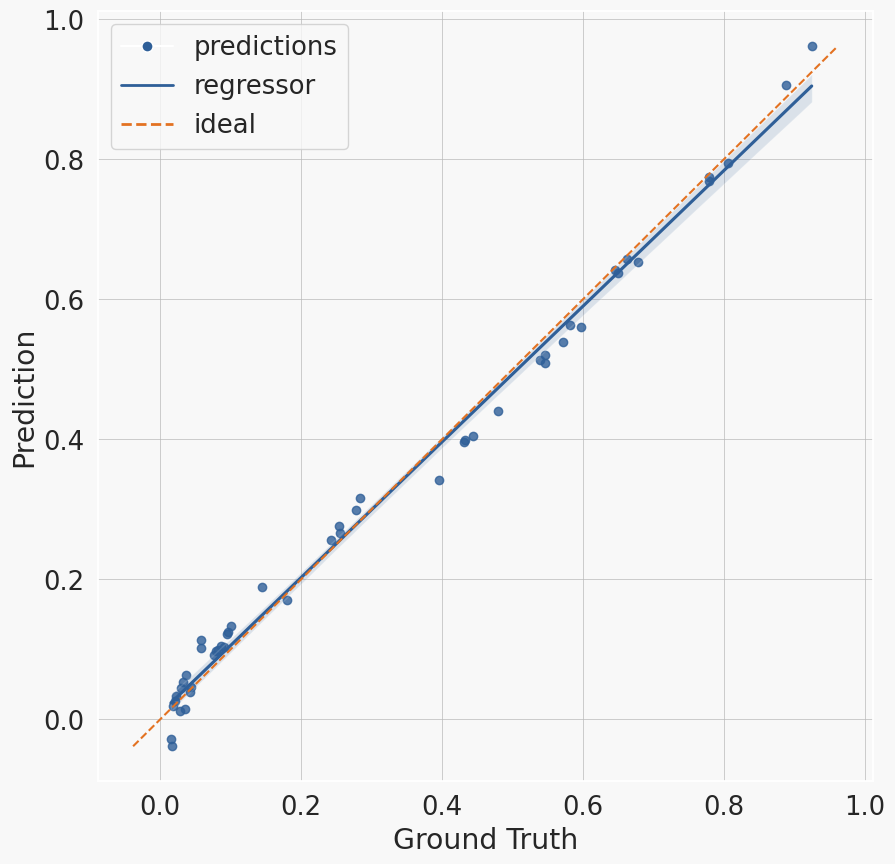}
            \caption[]%
            {{\small Sigmoid}}    
            \label{fig:layer-scatterplots sigmoid}
        \end{subfigure}
        \vskip\baselineskip
        \begin{subfigure}[b]{0.47\textwidth}   
            \centering 
            \includegraphics[width=\textwidth]{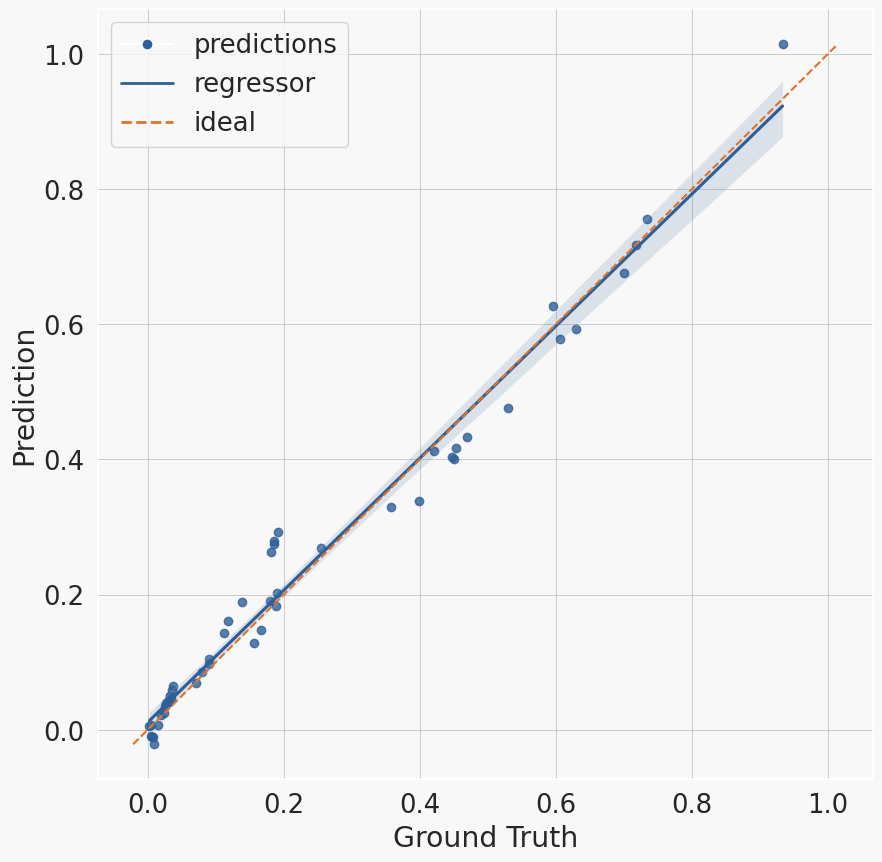}
            \caption[]%
            {{\small Tanh}}    
            \label{fig:layer-scatterplots tanh}
        \end{subfigure}
        \hfill
        \begin{subfigure}[b]{0.47\textwidth}   
            \centering 
            \includegraphics[width=\textwidth]{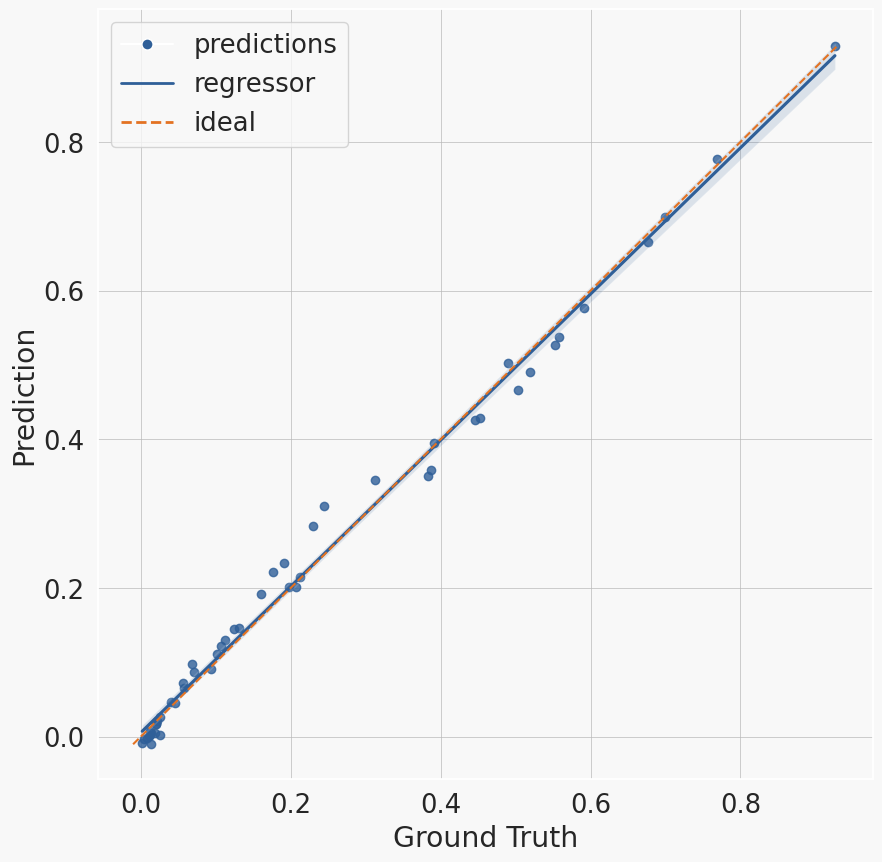}
            \caption[]%
            {{\small Softmax}}    
            \label{fig:layer-scatterplots softmax}
        \end{subfigure}
        \caption[Shows scatter-plots for the selected activation models (see Sec.~\ref{sec:results}), illustrating the accuracy of the predictions on the random test data; the x-axis represents the ground truth while the y-axis represents the predicted values, and the diagonal line symbolizes perfect predictions.]
        {Shows scatter-plots for the selected activation models (see Sec.~\ref{sec:results}), illustrating the accuracy of the predictions on the random test data; the x-axis represents the ground truth while the y-axis represents the predicted values, and the diagonal line symbolizes perfect predictions.}
        \label{fig:layer-scatterplots-activations}
\end{figure}
\begin{figure}[h]
    \centering
    \includegraphics[width=0.9\textwidth]{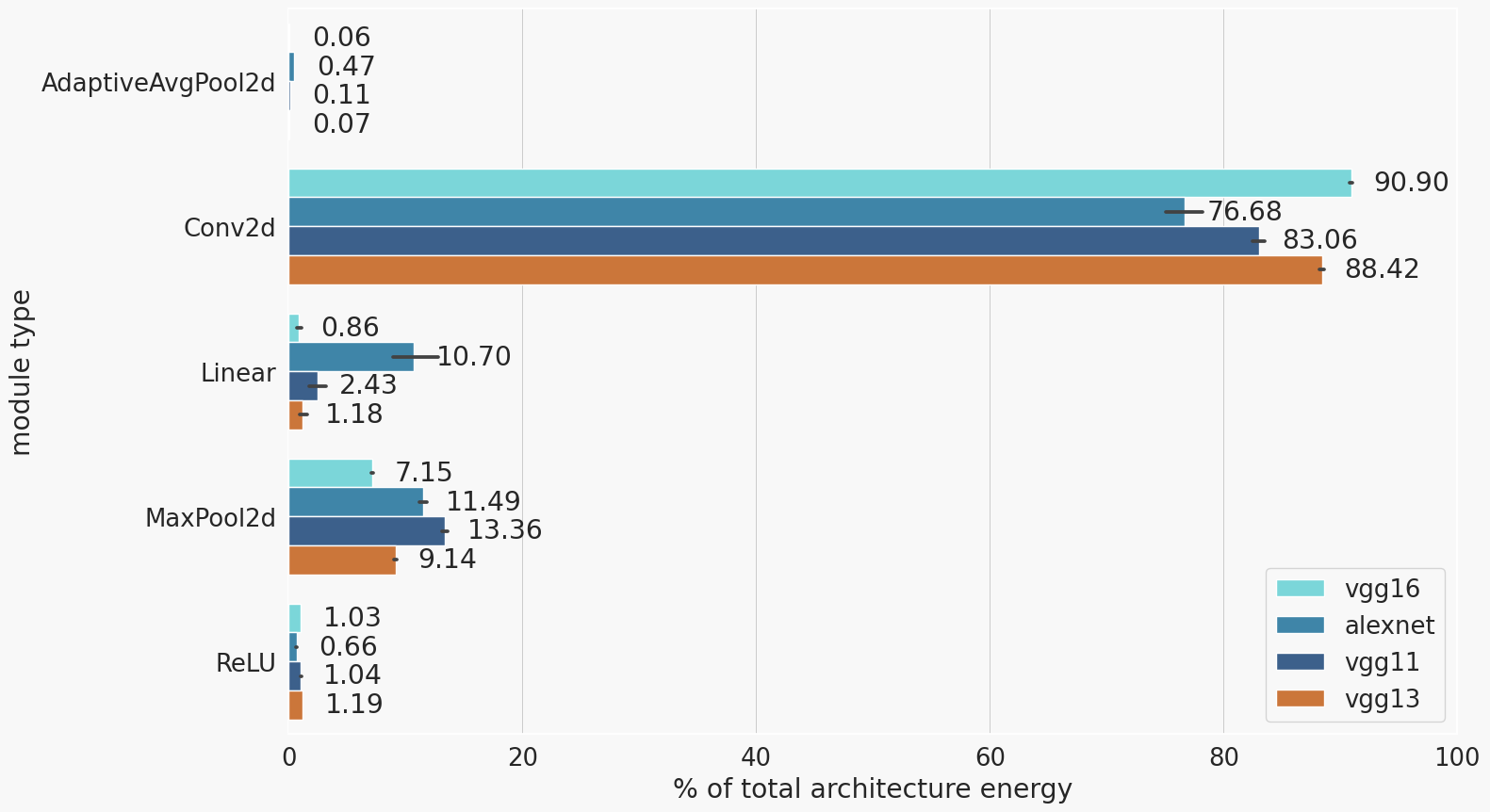}
    \caption{Shows the relative contributions of the individual layer types to the total energy consumption for AlexNet and VGG11/13/16; the black bars correspond to deviations caused by different batch sizes.}
    \label{fig:layer-wise-contribution}
\end{figure}
\begin{figure}
    \centering
    \includegraphics[width=0.9\textwidth]{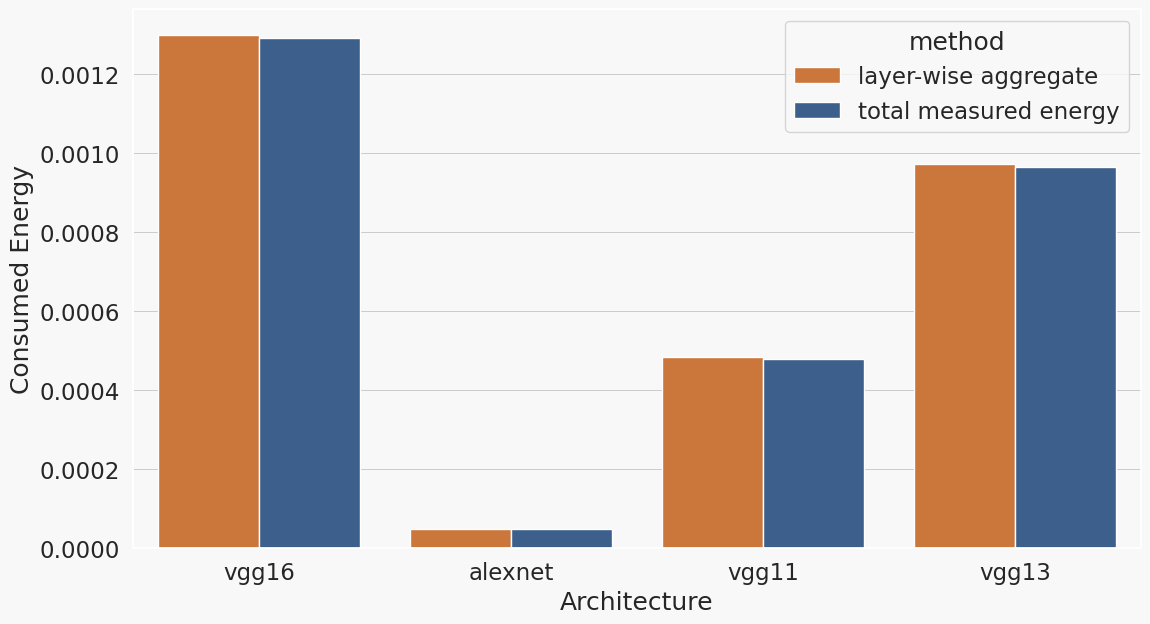}
    \caption{Compares the energy consumption of various architectures, when using two different measuring techniques; the ``total measured energy'' corresponds to the value obtained when measuring the architectures as singular units and the ``layer-wise aggregate'' corresponds to the sum of per layer measurements.}
    \label{fig:agg-vs-total-energy}
\end{figure}

\end{document}